\documentclass[conference]{IEEEtran}
\IEEEoverridecommandlockouts
\usepackage{cite}
\usepackage{amsmath,amssymb,amsfonts}
\usepackage{graphicx}
\usepackage{textcomp}
\usepackage{xcolor}
\usepackage{amsmath,amssymb,amsfonts}
\usepackage{graphicx}
\usepackage{textcomp}

\usepackage{tcolorbox}
\usepackage{subcaption}
\usepackage{algorithm}
\usepackage{algorithmicx}
\usepackage{algpseudocode}
\usepackage{comment}
\usepackage[misc]{ifsym}
\usepackage{soul}
\usepackage{booktabs} 
\usepackage{multirow}
\usepackage[colorinlistoftodos]{todonotes}
\usepackage{makecell}
\usepackage{enumitem}
\usepackage{flushend}
\usepackage{newfloat}
\usepackage{listings}
\usepackage{url}
\usepackage{ifsym}

\floatstyle{ruled}
\newfloat{listing}{tb}{lst}{}
\floatname{listing}{Listing}

\newcommand{\squishlist}{
	\begin{list}{$\bullet$}
		{ \setlength{\itemsep}{0pt}
			\setlength{\parsep}{3pt}
			\setlength{\topsep}{3pt}
			\setlength{\partopsep}{0pt}
			\setlength{\leftmargin}{1.5em}
			\setlength{\labelwidth}{1em}
			\setlength{\labelsep}{0.5em} } }
	
	\newcommand{\squishlisttwo}{
		\begin{list}{$\bullet$}
			{ \setlength{\itemsep}{0pt}
				\setlength{\parsep}{0pt}
				\setlength{\topsep}{0pt}
				\setlength{\partopsep}{0pt}
				\setlength{\leftmargin}{2em}
				\setlength{\labelwidth}{1.5em}
				\setlength{\labelsep}{0.5em} } }
		
\newcommand{\squishend}{
\end{list}  }
\usepackage{soul}
\usepackage{tikz}

\makeatletter
\newcommand{\linebreakand}{%
  \end{@IEEEauthorhalign}
  \hfill\mbox{}\par
  \mbox{}\hfill\begin{@IEEEauthorhalign}
}
\makeatother
    
\newcommand{\ours}{{\textsf{JAMES}}}
\newcommand{\jtm}{{\emph{JTN}}}
\newcommand{\ie}{{\textit i.e.}}
\newcommand{\eg}{{\textit e.g.}}

\def\BibTeX{{\rm B\kern-.05em{\sc i\kern-.025em b}\kern-.08em
    T\kern-.1667em\lower.7ex\hbox{E}\kern-.125emX}}

\begin{document}
\title{JAMES: Normalizing Job Titles with Multi-Aspect Graph Embeddings and Reasoning}



\author{
Michiharu Yamashita$^{1}$,
Jia Tracy Shen$^{1}$,
Thanh Tran$^{2}$,
Hamoon Ekhtiari$^{3}$,
and Dongwon Lee$^{1}$ \\
$^{1}$The Pennsylvania State University, University Park, PA, USA \\
$^{2}$Amazon, Cambridge, MA, USA \> 
$^{3}$FutureFit AI, Toronto, Canada \\
\{michiharu, jqs5443\}@psu.edu, tdt@amazon.com, hamoon@futurefit.ai, dongwon@psu.edu
}

\maketitle
\begin{abstract}
In online job marketplaces, it is important to establish a well-defined job title taxonomy for various downstream tasks (\eg, job recommendation, users' career analysis, and turnover prediction). \emph{Job Title Normalization} ({\jtm}) is such a cleaning step to classify user-created non-standard job titles into normalized ones. 
However, solving the {\jtm} problem is non-trivial with challenges: 
(1) semantic similarity of different job titles, 
(2) non-normalized user-created job titles, 
and (3) large-scale and long-tailed job titles in real-world applications. 
To this end, we propose a novel solution, named {\ours}, that constructs three unique embeddings (\ie, \emph{graph}, \emph{contextual}, and \emph{syntactic}) of a target job title to effectively capture its various traits. We further propose a multi-aspect co-attention mechanism to attentively combine these embeddings, and employ neural logical reasoning representations to collaboratively estimate similarities between messy job titles and normalized job titles in a reasoning space. 
To evaluate {\ours}, we conduct comprehensive experiments against {\em ten} competing models on a large-scale real-world dataset with over 350,000 job titles. Our experimental results show that {\ours} significantly outperforms the best baseline by 10.06\% in Precision@10 and by 17.52\% in NDCG@10, respectively. 
To further facilitate the acquisition of normalized job titles for job-domain applications, our {\ours} API is available at: \url{https://tinyurl.com/james-job-title-mapping}.
\end{abstract}


\section{Introduction}
\begin{figure*}[t]
  \centering
  \includegraphics[width=0.8\linewidth]{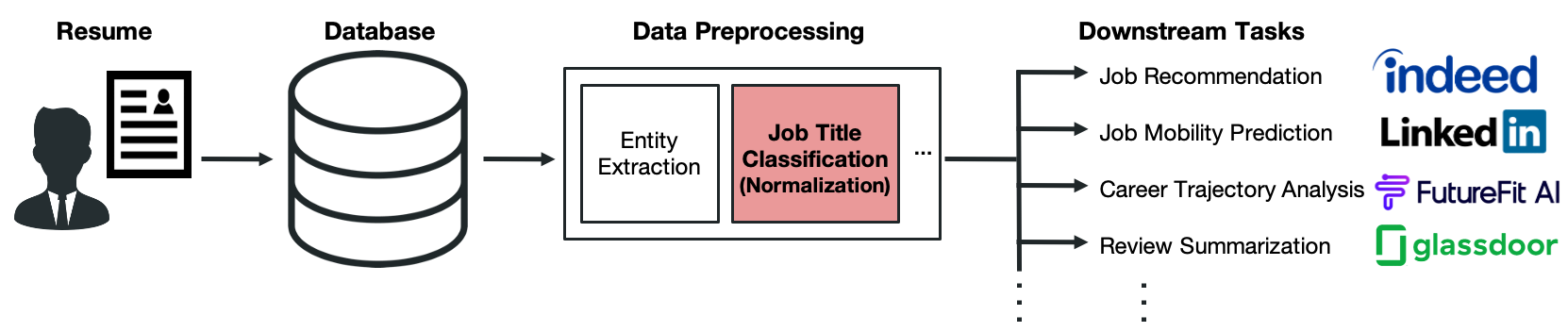}
  \vspace{-5pt}
  \caption{Workflow of job-domain applications}
  \label{fig:downstream_tasks}
  \vspace{-10pt}
\end{figure*}

\noindent{\bf Background.} 
The recent proliferation of technology has witnessed an increasing popularity of online professional platforms. These online job marketplaces connect job seekers and companies to find the best match for each other. For example, LinkedIn and Indeed, two of the largest jobs marketplace platforms, have more than 930 million users\footnote{https://about.linkedin.com/} and 245 million resumes\footnote{https://www.indeed.com/about}, respectively. 
The vast amount of data available on job marketplaces, including resumes from job seekers and job postings from companies, has spurred companies involved in workforce development, talent intelligence, recruitment, and job search engines to utilize AI techniques to enhance their applications (\eg, job recommendation \cite{kenthapadi2017personalized, paparrizos2011machine}, next career prediction \cite{meng2019hierarchical, xu2018dynamic}, and career analysis \cite{li2017prospecting}). These AI-powered tools enable job seekers in finding their ideal jobs and companies in recruiting talents that match their roles. 
However, the workflow of such job-domain applications involves a critical step, as illustrated in Figure \ref{fig:downstream_tasks}. Before building models for downstream tasks, the various entities found in raw data, especially {\em job titles}, must be sorted, consolidated, and normalized. For instance, a position called ``systems engineer" in a company $A$ and another position called ``application programmer" in a company $B$ may refer to the same job. Normalizing these two job titles into  ``software developer'' (or noting their compatibility in context) is crucial for job recommendations, career trajectory analysis, and search result expansion. Therefore, the research question (RQ) we investigate is: \emph{How can job titles be automatically normalized}? In particular, we aim to answer this RQ via the framing of the Job Title Normalization ({\jtm}) (to be defined in Section 3.1). 

\vspace{0.1mm}
\noindent{\bf Challenges.} 
Although the {\jtm} problem appears simple in nature, addressing it in practice poses several challenges. 
First, job titles often bear a semantic closeness to one another that is contingent upon the required skill sets and companies' own definitions. For example, the job title ``data scientist'' is prevalent today, and comprises skill sets such as mathematical modeling, statistics, and coding. However, this role can be related to ``business analyst'' or ``data analyst'' in some companies, and to ``product scientist'' in others. Thus, comparing job titles alone is inadequate for solving {\jtm}, and it is necessary to represent them in a semantic space to ensure accurate calibration. 
Second, job titles collected from users' resumes are often untidy due to non-standard naming conventions and auxiliaries. The job title ``software developer" in one resume can be written as ``SDE" in another. Moreover, creative job titles such as ``data geek" or ``strategic futurist" that individuals may list on their resumes do not necessarily appear in an industry-wide job title taxonomy. 
Third, while an industry job taxonomy contains only a few hundred to a few thousand job titles, the number of job titles encountered on job marketplace platforms is orders of magnitude larger. Nevertheless, existing solutions have only employed either small-scale datasets or company-created datasets (as opposed to user-written), in which {\jtm} was addressed through manual labeling/cleaning or text normalization procedures. For instance, Zhang et al. \cite{zhang2020large} employed a dataset with only 26 unique job titles for similar expertise job matching, while Dave et al. \cite{dave2018combined} used a dataset of 4,325 unique job titles for job and skill recommendation tasks. More recently Li et al. \cite{li2020deep} developed a job title taxonomy containing 30,000 entries on LinkedIn for a job understanding task. However, the challenges inherent in {\jtm} for the vast number of job titles still remain inadequately addressed. 

\noindent{\bf Ideas.} 
To address the aforementioned challenges in {\jtm}, we propose {\ours} (\underline{J}ob title m\underline{A}pping with \underline{M}ulti-aspect \underline{E}mbeddings and rea\underline{S}oning), and demonstrates its effectiveness using a real-world career dataset containing more than 350,000 job titles. Specifically, {\ours} considers three unique multi-aspect (\ie, graph, contextual, and syntactic) embeddings for candidate job titles. 
First, we establish a graph embedding to represent the latent {\em topological} job title similarity based on users' job transitions, exploiting the fact that users typically switch to similar positions or titles (\ie, changing from ``data scientist" to ``chef" is highly unlikely although possible). We use a hyperbolic graph embedding for the latent knowledge dependencies in a job title hierarchy, as it outperforms Euclidean graph embeddings on hierarchical structure datasets \cite{nickel2017poincare, ganea2018hyperbolic}. In addition, hyperbolic graph embeddings help mitigate the problem of incomplete and inconsistent job transition patterns by providing a smaller distortion and an exponential expansion of nodes \cite{chami2019hyperbolic, ganea2018hyperbolic}.
Second, we leverage a pretrained BERT embedding to account for the {\em contextual} similarity between two candidate job titles, which can identify contextually-related job titles, as language models can measure the contextual and semantic distance. 
Third, we create a {\em syntactic} embedding to capture the string-to-string similarity between two input job titles, allowing for the detection of misspelled (\eg, ``electric engieer'') and user-created (\eg, ``cool data scientist'') job titles. 
Also, we design a neural collaborative reasoning \cite{chen2021neural} that takes multi-aspect embeddings as input and produces reasoning-based multi-aspect embeddings to mitigate uncertainty among standard job titles, covering job titles that are not accurately captured by either contextual or syntactic embeddings. 
After building multi-aspect embeddings using our large-scale resume dataset, we develop a multi-aspect co-attention mechanism that considers all three multi-aspect embeddings concurrently. 

\vspace{1mm}
\noindent{\bf Contributions.} 
Our contributions are as follows:
\vspace{-2pt}
\begin{itemize}
\item We use a large-scale, real-world, and user-generated dataset from a career platform (FutureFit AI), which comprises over 350,000 unique job titles, for the job-domain specific preprocessing task, \emph{Job Title Normalization} ({\jtm}). 
\item To solve the {\jtm} task, we propose a novel model, {\ours}, that employs multi-aspect embeddings and reasoning representations accounting for \emph{graph}, \emph{contextual}, and \emph{syntactic} embeddings.
\item We conduct extensive experiments and demonstrate the effectiveness of {\ours} against {\em ten} competing baseline models. {\ours} significantly outperforms the best baseline by 10.06\% in Precision@10 and by 17.52\% in NDCG@10, respectively. We also apply {\ours} to other downstream tasks and report the findings and further implications.
\item We develop and release {\ours} API publicly, allowing for the acquisition of normalized job titles from job title entities.
\end{itemize}

\section{Related Work}
\subsection{Job Title Classification}
Previously, {\jtm} was often overlooked and just addressed through manual labeling or simple data preprocessing. However, there have been several prior works that attempt to solve it as a task of job title classification \cite{giabelli2020neo, boselli2018wolmis}. Wang et al. \cite{wang2019deepcarotene} proposed a CNN-based approach that developed text vectors using a job description dataset, while Zhu et al. \cite{zhu2016semantic} built a KNN model using Word2Vec. While such methods using job descriptions can be helpful, in the real world, it is often difficult to obtain access to all companies' job description datasets, and the applicability of such methods to user-generated job titles extracted from resumes is not well understood. Therefore, our work aims to develop a practical solution applicable to a user-generated dataset (\ie, resumes).

As a job entity benchmarking, Luo et al. \cite{luo2019learning} created a job transition graph using Random Walk-based vectors, and indicated the potential of job graph embedding. Zhang et al. \cite{zhang2019job2vec} proposed Job2Vec as a job title benchmarking tool based on job records. However, both works were only validated in link and/or node prediction and not specifically designed for {\jtm}, resulting in uncertainty regarding their applicability to {\jtm} and normalization. Moreover, Job2Vec aimed to link job titles of the same expertise level to calibrate salaries for recruiters \cite{zhang2019job2vec}. While these benchmarks could be used for job title clustering, they manually filtered out low-frequency words in job titles as a data preprocessing step, and limited their dataset to the IT and finance domains, which restricts the generalizability of their representations to real-world scenarios. Additionally, only 15 well-known companies such as Google and Microsoft were chosen in their IT dataset, which may not reflect a practical scenario where companies or individuals want to map all job titles from users' resumes into normalized ones.
In contrast, our study focuses on a realistic and large-scale setting for {\jtm}, utilizing a dataset from 165,086 unique companies across all sectors. Our approach differs from previous works as we employ contextual embeddings to capture the potential meaning of words, and syntactic embeddings to detect misspelled and user-created words, addressing issues not considered in prior works. Furthermore, we use reasoning to obtain more robust representations.

\subsection{Representation Learning in Job Domain}
Liu et al. \cite{liu2016fortune} conducted career path prediction using multiple social media. 
For job skill representation, Shi et al. \cite{shi2020salience} developed ``Job2Skills", a market-aware skill extraction system, which considers the salient level of a skill and extracts important skill entities from job postings and target members using multi-resolution. 
Qin et al. \cite{qin2018enhancing} developed a person-job fit model that applied a word-level semantic representation for both job requirements and job seekers’ experiences based on RNN. 
Yamashita et al. \cite{yamashita2022looking} proposed a long-term career path prediction from large-scale resumes with multiple embeddings. 
While these studies relied on job-domain entities, our {\ours} can be applied to such job-domain applications to normalize job titles. 

\subsection{Hyperbolic Machine Learning}
Hyperbolic geometry is a non-Euclidean geometry that focuses on spaces of constant negative Gaussian curvature. Hyperbolic space has been used to develop embedding and machine learning models for hierarchical and graph structures, due to its benefits such as embedding on smaller dimensions \cite{ganea2018hyperbolic}. Poincare embedding, proposed by Nickel et al. \cite{nickel2017poincare}, enables hierarchical data to be represented better than Euclidean embeddings. Chami et al. \cite{chami2019hyperbolic} developed a hyperbolic technique for graph convolutional networks. In the case of career trajectory datasets, they can be represented as graphs, as done by Zhang et al. \cite{zhang2020large}. However, to the best of our knowledge, our work is the first to apply hyperbolic geometry to career transition data. Since hyperbolic embeddings work well on tree-structured datasets, we consider hyperbolic embeddings to be effective for representing latent knowledge dependencies in job titles, which are often hierarchical (\eg, junior, senior, VP). Hence, we incorporate hyperbolic embeddings into our model and compare the baselines.


\section{Preliminary}
\newcommand{\topoembed}{$\boldsymbol{X_{h}}\;$}
\newcommand{\semanticembed}{$\boldsymbol{X_{b}}\;$}
\newcommand{\bertembed}{\semanticembed}
\newcommand{\stringembed}{$\boldsymbol{X_{s}}\;$}
\newcommand{\affgraphbert}{$ A^{(hb)}\;$}
\newcommand{\affgraphsim}{$A^{(hs)}\;$}
\newcommand{\affbertsim}{$A^{(bs)}\;$}
\newcommand{\weighttopobert}{$ W^{(hb)}\;$}
\newcommand{\weighttoposim}{$W^{(hs)}\;$}
\newcommand{\weightbertsim}{$W^{(bs)}\;$}
\newcommand{\attentiongraph}{$K_h\;$}
\newcommand{\attentionbert}{$K_b\;$}
\newcommand{\attentionstring}{$K_s\;$}
\newcommand{\outputattgraph}{$\boldsymbol{\hat{X}_{h}}\;$}
\newcommand{\outputattbert}{$\boldsymbol{\hat{X}_{b}}\;$}
\newcommand{\outputattstring}{$\boldsymbol{\hat{X}_{s}}\;$}
\newcommand{\outputreasonbert}{$\boldsymbol{{X}^\prime_{b}}\;$}
\newcommand{\outputreasonstring}{$\boldsymbol{{X}^\prime_{s}}\;$}
\newcommand{\job}{j}
\newcommand{\standardjob}{v}
\newcommand{\standardjobset}{$\mathcal{Y}$}

\begin{table}[t!]
   \centering
      \caption{Definition of our notations in this paper}
      \vspace{-3pt}
      \label{tab:symbol}
      \centering
        \begin{tabular}{c|p{7cm}}\hline
            Notation & Definition\\\hline
            $\job$& job title from resume\\
            $\mathcal{X}$ & a set of job titles from all resumes \\
            $\standardjob$& normalized job title from the ground truth\\
            \standardjobset & a set of normalized job titles from the ground truth \\
            \topoembed & hyperbolic graph embeddings\\
            \bertembed & BERT embeddings\\
            \stringembed& syntactic (string similarity) embeddings\\
            \affgraphbert& affinity matrices between \topoembed and \semanticembed\\
            \affgraphsim& affinity matrices between \topoembed and \stringembed \\
            \affbertsim& affinity matrices between \semanticembed and \stringembed \\
            \weighttopobert& learnable weights between \topoembed and \semanticembed \\
            \weighttoposim& learnable weights between \topoembed and \stringembed \\
            \weightbertsim& learnable weights between \semanticembed and \stringembed\\
            \attentiongraph& attention graph for \topoembed, acknowledging supports from the other embedding views \semanticembed and \stringembed through \affgraphbert and \affgraphsim \\
            \attentionbert& attention graph for \semanticembed, acknowledging supports from the other embedding views \topoembed and \stringembed through \topoembed and \affgraphsim \\
            \attentionstring& attention graph for \stringembed, acknowledging supports from the other embedding views \semanticembed and \topoembed through \affgraphbert and \topoembed\\
            \outputattgraph & co-attentive multi-view embeddings from \topoembed\\
            \outputattbert & co-attentive multi-view embeddings from \semanticembed\\
            \outputattstring & co-attentive multi-view embeddings from \stringembed\\
            \outputreasonbert & reasoning-based representation from \topoembed\\
            \outputreasonstring & reasoning-based representation from \stringembed\\\hline
        \end{tabular}
    \vspace{-5pt}
\end{table}

Major notations used throughout the paper are summarized in Table \ref{tab:symbol}. 
In this section, we describe a definition of the problem, Job Title Normalization ({\jtm}), and our dataset. 

\subsection{Problem Definition}
We formally define the {\jtm} task as follows:
\begin{tcolorbox}
\textbf{Job Title Normalization ({\jtm})}: Given a set of job titles $\mathcal{X}$ and a set of predefined and normalized job titles in a job taxonomy $\mathcal{Y}$, the \emph{Job Title Normalization} task aims to build a function $f(\cdot)$ that produces matching probabilities from each job title $\job_i \in \mathcal{X}$ to each normalized job title $\standardjob_k \in \mathcal{Y}$. 
\end{tcolorbox}

Note that during the training of the mapping function $f(\cdot)$, we consider the {\jtm} task as a multi-class classification task. During the inference, with each $\job_i \in \mathcal{X}$, we take the output probability distribution over all normalized job titles $\standardjob_k \in \mathcal{Y}$ as the ranking scores to output \emph{top-k} most similar job titles $\standardjob_k \in \mathcal{Y}$.

\subsection{Dataset}

We obtained the dataset from a popular career platform
FutureFit AI\footnote{https://www.futurefit.ai/}, 
which partners globally with other companies and governments to assist employees in navigating career transitions. We randomly selected over 400,000 resumes from the platform, which had at least five valid working experiences in the United States (\ie, a path with five nodes in a job transition graph). This was done to ensure that the job transition graph was meaningful, while also being reasonably large. Table \ref{tab:stat} presents a summary of our dataset information, including the average length of words and characters in job titles.

Our dataset contains 354,168 unique job titles from 165,086 unique companies, and more than 2.7 million job transition trajectories. To solve the {\jtm} task, we only extract the job seeker's basic information from their resume, such as company ID, job title, start and end dates of working, while ensuring that all private employee information is anonymized. On average, job titles have 4.15 words and 28.9 characters. The top five most frequent job titles are sales associate (0.33\%), research assistant (0.23\%), administrative assistant (0.23\%), project manager (0.19\%), and CEO (0.18\%). 
Our dataset will be available upon request. 

In comparison to previous works \cite{zhang2020large, dave2018combined, li2017nemo, zhang2019job2vec}, our dataset has a significantly larger number of job titles. Specifically, our dataset contains over 11,500 times more job titles than \cite{zhang2020large}, over 69 times more job titles than \cite{dave2018combined}, over 30 times more job titles than \cite{li2017nemo}, and four times more job titles than the Job2Vec (Finance) dataset \cite{zhang2019job2vec}. 
To create our ground truth dataset, we perform an exact search to match job titles in our dataset with the European Skills/Competences, qualifications, and Occupations (ESCO) taxonomy\footnote{https://ec.europa.eu/esco/portal/escopedia/ESCO}, which provides a hierarchical structure of job titles and normalized job titles as job groups. For example, ``software developers" consists of ``application developer", ``software engineer", ``software architect", etc. We remove proper nouns from the job titles, and use the matched job titles as the ground truth labels for our experiments.

\begin{table}[t]
   \centering
      \caption{Our large-scale resume dataset}
      \label{tab:stat}
      \centering
      \resizebox{0.65\linewidth}{!}{
        \begin{tabular}{l|r}\hline
            \# of Resumes&401,253\\\hline
            \# of Job Titles&354,168\\\hline
            \# of Companies&165,086\\\hline
            \# of Transitions&2,738,403\\\hline
            Average length of words&4.15\\\hline
            Average length of characters&28.9\\\hline
        \end{tabular}
        }
\end{table}


\section{Our Proposed Model: {\ours}}
\begin{figure}[tb]
  \centering
  \includegraphics[width=0.9\linewidth]{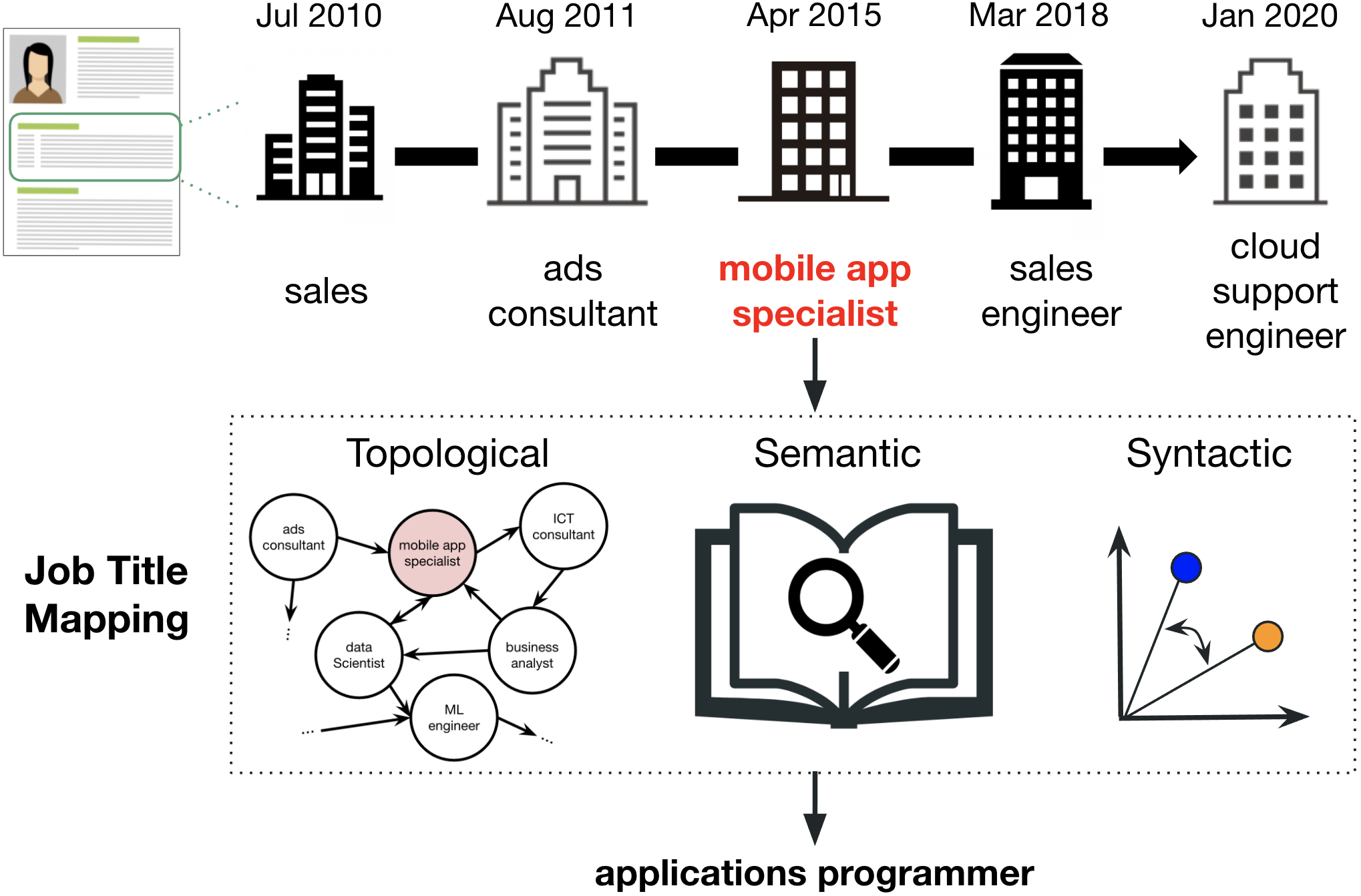}
  \caption{Toy example of {\ours} in \emph{Job Title Normalization}.}
  \label{fig:toy}
  \vspace{-10pt}
\end{figure}

\begin{figure}[t]
  \centering
  \includegraphics[width=\linewidth]{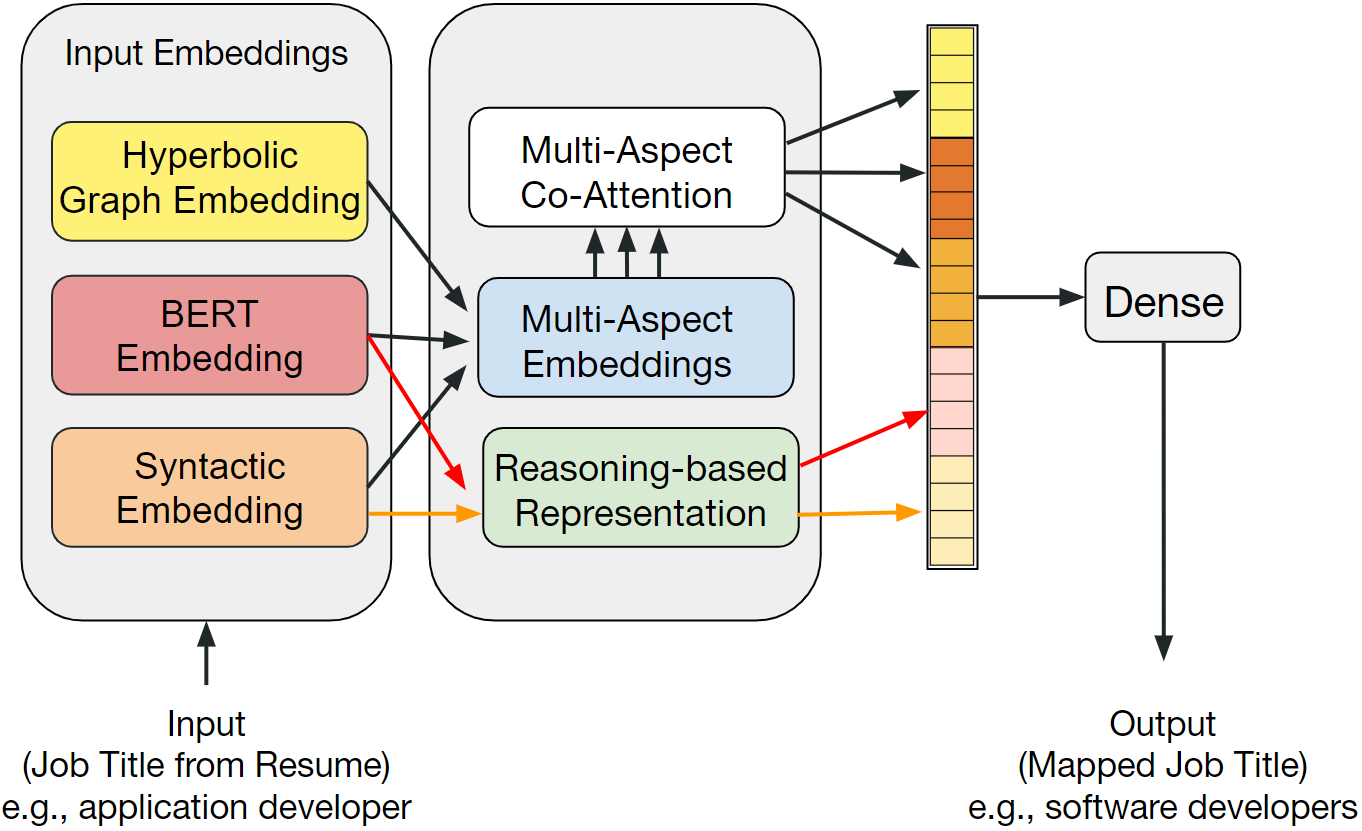}
  \caption{Model overview of {\ours}.
  \label{fig:model_overview}
  }
  \vspace{-5pt}
\end{figure}

In this section, we describe our job title normalization model, {\ours}. The main idea of {\ours} is to learn multi-aspect representations of an input job title and produce its corresponding \emph{top-n} normalized standard job title mappings that are predefined in a standard job title taxonomy. Figure \ref{fig:toy} shows a toy example of how {\ours} works. First, the job titles of a resume are extracted. Next, {\ours} learns the multi-aspect embeddings, including graph, semantic, and syntactic embeddings, for each job title. In this example, the input job title is ``mobile app speciallist'', and {\ours} utilizes the multi-aspect embeddings to predict the matching standard job title, resulting in ``applications programmer''.

Figure \ref{fig:model_overview} provides a more detailed overview of {\ours}. To learn the graph embeddings of the input job title, {\ours} employ the state-of-the-art hyperbolic graph representation learning. To learn the semantic embeddings of the input job title, {\ours} use the well-known pretrained BERT. To obtain the syntactic embeddings of the input job title, {\ours} encodes a dense embedding vector with a size equal to the number of standard job titles. Each element in the vector represents the string-based similarity score between the input job title and the corresponding standard job title. Next, we propose a multi-aspect co-attention mechanism that assigns attention scores to the three multi-aspect embeddings. We also introduce a reasoning-based module in {\ours} that collaboratively reasons the multi-aspect embeddings in a reasoning space. Finally, {\ours} fuses all the output embeddings to produce the \emph{top-n} mapping normalized job titles for the input job title as outputs. We provide a detailed description of {\ours} in the following subsections.

\subsection{Multi-Aspect Embeddings}

\subsubsection{Hyperbolic Graph Embedding}
\begin{figure}[tb]
  \centering
  \includegraphics[width=0.9\linewidth]{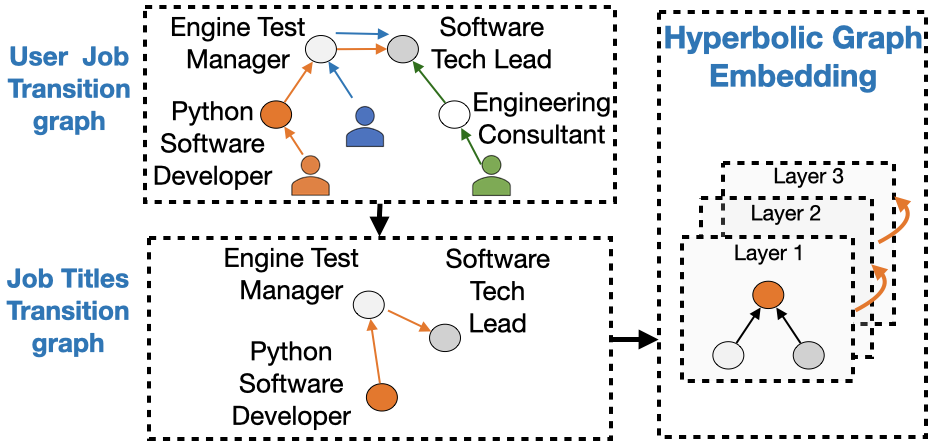}
  \caption{Architecture for hyperbolic graph embedding.}
  \label{fig:hge}
  \vspace{-10pt}
\end{figure}

\begin{figure}[t]
 \begin{subfigure}{0.4\linewidth}
 \centering
    \includegraphics[width=0.8\linewidth]{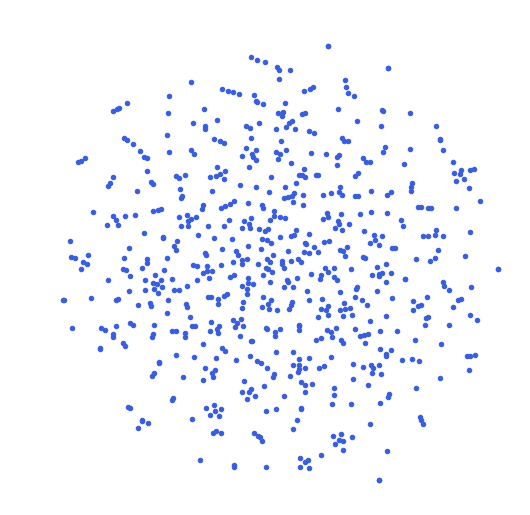}
    \caption{Euclidean embedding}
  \end{subfigure}
  \hspace*{\fill} 
  \begin{subfigure}{0.4\linewidth}
  \centering
    \includegraphics[width=0.8\linewidth]{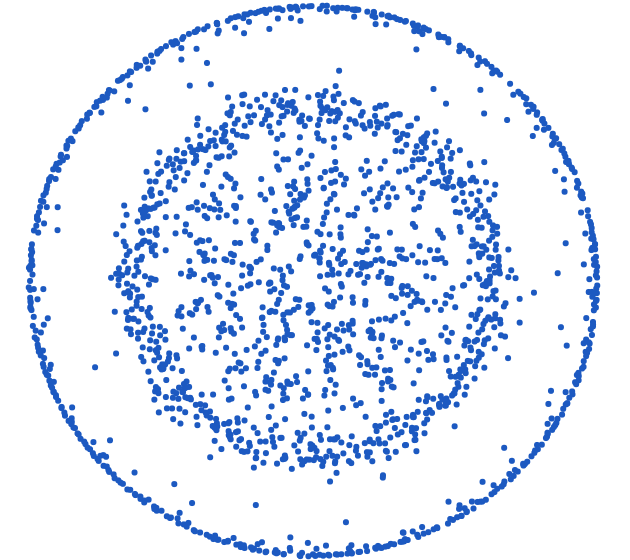}
    \caption{Hyperbolic embedding}
  \end{subfigure}
 \caption{Example visualizations for the Euclidean and Hyperbolic embeddings from job transition graph.}
 \label{fig:poincare}
 \vspace{-5pt}
\end{figure}

To construct a hyperbolic graph embedding that captures the topological features of our career trajectory dataset, we first create a job transition graph, as illustrated in Figure \ref{fig:hge}. We define nodes as job titles and links as the transitions between the job titles, where each link is directed and asymmetric. For instance, if a person changes their job title from ``Software Engineer" (SWE) to ``Machine Learning Engineer" (MLE), the graph has a directed link from SWE to MLE. 

The job transition graph is defined as $G = (V, E, W)$, where $V$ is the set of job titles (\ie, nodes), $E$ is the set of job transitions (\ie, links in the graph), and $W$ is the set of link weights. The job transition weight $W_{i,j}$ is formulated as $W_{i,j} = e_{i, j} / \sum_{i=1}^{n} \sum_{j=1}^{n} e_{i, j}$, where $e_{i, j}$ is the number of transitions from node $v_i$ to node $v_j$. Based on all job transitions, we construct a graph and derive graph embeddings. To build the hyperbolic graph, we consider a head node (\ie, more recent user's career) as the \textit{parent} and a tail node (\ie, the previous career) as the \textit{child}, assuming that the most recent job title contains all the requirements and skill sets from the previous job titles. We then embed the nodes in hyperbolic space using Poincare embedding \cite{nickel2017poincare} as a hyperbolic embedding and train a Poincare ball model from the relations of nodes in the graph.

Since the Poincare ball is a Riemannian manifold, the Riemannian metric tensor is represented in the $d$-dimentional ball $B^d = \{x \in \mathbb{R}^d | ||x|| < 1\}$, where $||x||$ is the
Euclidean norm. Then, the Riemannian metric tensor $r_x$ is defined as:
\begin{equation}
    r_x = \Bigl(\frac{2}{1 - ||x||^2} \Bigl)^2 r^E
\end{equation}
where $x \in B^d$ and $r^E$ is the Euclidean metric tensor. Then, the distance of two points $a, b \in B^d$ is defined as:
\begin{equation}
    d(a,b) = arcosh \Bigl(1 + 2 \frac{||a-b||^2}{(1-||a||^2)(1-||b||^2)} \Bigl)
\end{equation}
Based on these metrics, we construct the Poincare embedding on the Poincare ball \cite{nickel2017poincare} with the input of the parent-child pair dataset and obtain the $m$-dimensional embedding. Figure \ref{fig:poincare} provides a visualization comparison between Euclidean and Poincare embeddings in a 2-dimensional ball, where the hyperbolic embedding exhibits a hierarchy of dots, while each dot in the Euclidean embedding is scattered disorderly. We output \topoembed as the hyperbolic graph embedding of each input job title.

\subsubsection{BERT Embedding}
\begin{figure}[t]
  \centering
  \includegraphics[width=\linewidth]{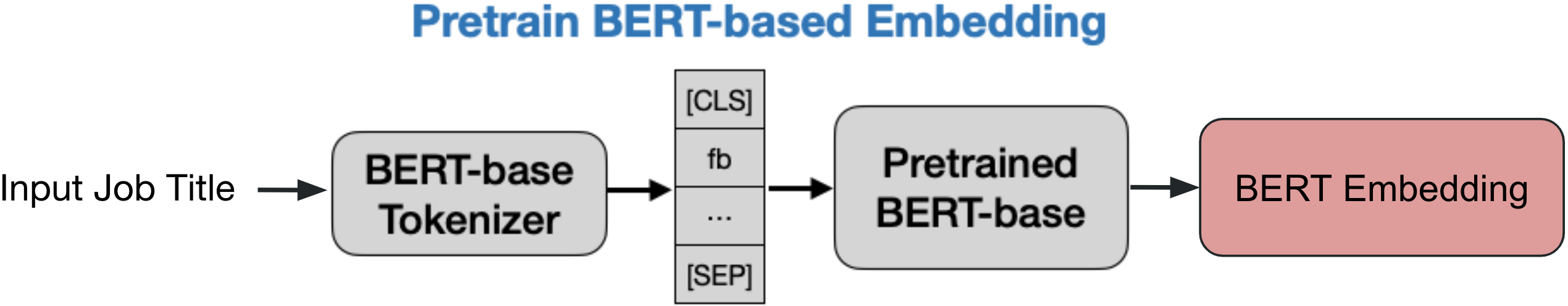}
  \caption{Learning semantic embeddings of the input job tile via the pretrained uncased BERT-base. }
  \label{fig:bert}
  \vspace{-10pt}
\end{figure}

To address the issue of different job titles referring to the same position (\eg, ``Data Analyst'' vs ``Data Scientist''), we learn the semantic embeddings of the input job titles using the pre-trained BERT \cite{devlin2018bert}. Specifically, we use the pre-trained DistilRoBERTa on SBERT \cite{reimers-2019-sentence-bert} due to its efficiency and effectiveness.

The architecture for the BERT embedding is illustrated in Figure \ref{fig:bert}. For each input job title, we first tokenize it into wordpieces using the BERT-base uncased tokenizer. Then, we use the pretrained BERT-base uncased embeddings to obtain the embeddings \bertembed of the [CLS] token as the final representations of the input job title.

\subsubsection{Syntactic Embedding}
\begin{figure}[t]
  \centering
  \includegraphics[width=0.8\linewidth]{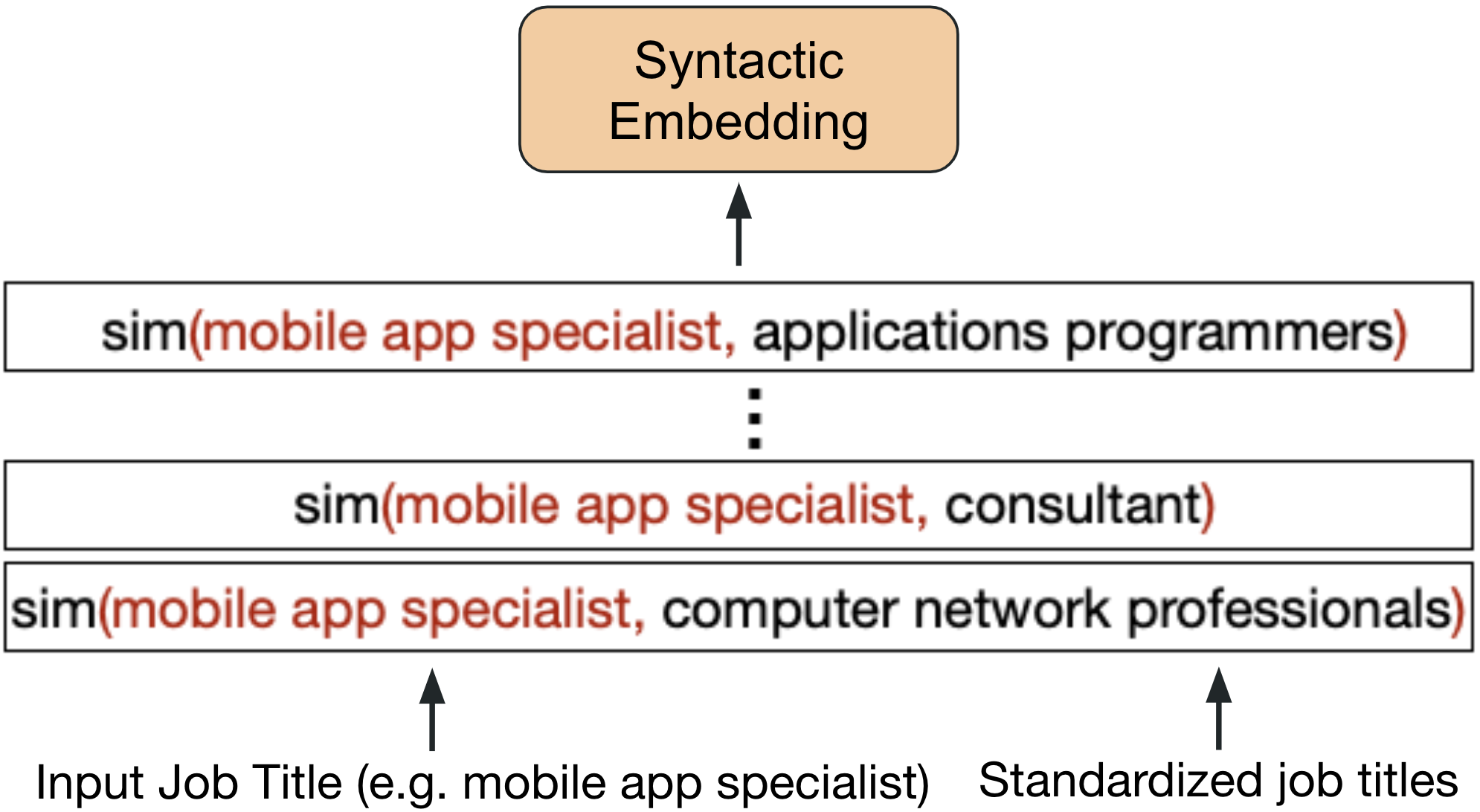}
  \caption{Architecture for syntactic (string similarity) embedding. Given the input job title ``\emph{mobile app specialist}'', we represent it by an embedding vector \stringembed 
  $\in \mathcal{R}^{\vert\mathcal{Y}\vert}$, where \stringembed[k] = \emph{cosine-sim}(``\emph{mobile app specialist}'', $\standardjob_k$), with $\standardjob_k \in$ \standardjobset.}
  \label{fig:string}
  \vspace{-10pt}
\end{figure}

To capture the syntactic representation of job titles, we use cosine similarity to score the string similarity between an input job title and all of the normalized job titles in the job taxonomy (\ie, ESCO). Figure \ref{fig:string} shows the architecture for this embedding. We calculate all pairs between job titles and their parent job titles, and define the similarity matrix as the syntactic representation. 

For a given set of 
$\mathcal{X}$ input job titles and $\mathcal{Y}$ predefined normalized job titles in the job taxonomy, the syntactic embedding of a job title $s$ $\in$ $\mathcal{X}$ is a vector $\boldsymbol{X_{s}}$ of $\mathcal{Y}$ dimensions, where each dimension indicates the string-string cosine similarity with a character-level comparison between $s$ and each corresponding normalized job title in $\mathcal{Y}$. For example, $\boldsymbol{X_{s}} [0] = \text{cosine-sim}(s, v_0)$, $\boldsymbol{X_{s}} [1] = \text{cosine-sim}(s, v_1)$, ..., $\boldsymbol{X_{s}} [\vert \mathcal{Y} \vert - 1] = \text{cosine-sim}(s, v_{\vert \mathcal{Y} \vert - 1})$. We define the resulting syntactic embeddings as \stringembed. 

\subsection{Multi-Aspect Co-Attention}
In the previous sections, we extract three discrete embeddings for an input job title: (i) hyperbolic graph embeddings \topoembed, (ii) BERT embeddings \bertembed, and (iii) syntactic embeddings \stringembed. However, the embeddings are learned separately and may have redundant features. To address this issue, we aim to learn multi-aspect embeddings that incorporate all three embeddings and are attentive to each other. For this purpose, a traditional method is to weigh each embedding view by hierarchical attention \cite{yang2016hierarchical}, where the \topoembed can be used as query, \bertembed can be used as key/value. Then the \topoembed and the \textit{attentive} \bertembed can be combined as a query, and \stringembed can be used as key/value. As the hierarchical attention is performed sequentially and is not practical for large-scale datasets with millions of job titles. Therefore, we extend the traditional co-attention mechanism \cite{lu2016hierarchical} which takes only two input sources, and propose a multi-aspect co-attention mechanism that can work for $p$ inputs (\ie, $p \ge 2$). In this sense, our multi-aspect co-attention mechanism uses $k-1$ views to guide the attention weights for the left-over view in parallel. 

\begin{figure}[t]
  \centering
  \includegraphics[width=0.9\linewidth]{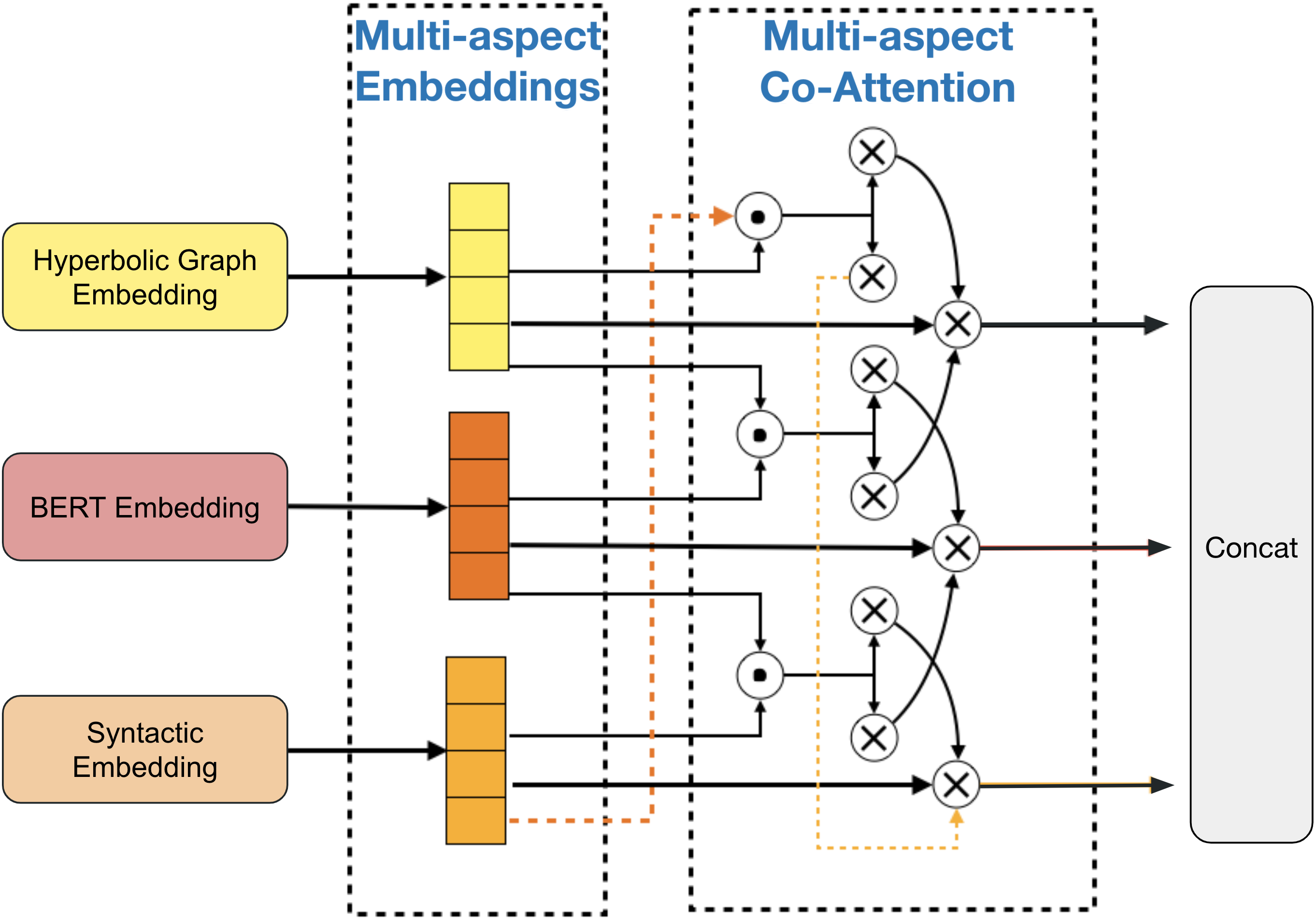}
  \caption{Architecture for Multi-view Co-attention.
  }
  \label{fig:coatt}
  \vspace{-10pt}
\end{figure}

Figure \ref{fig:coatt} shows the architecture for our multi-aspect co-attention. 
We start by computing three affinity matrices for three pairs of two embedding views: \affgraphbert between \topoembed and \semanticembed, \affgraphsim between \topoembed and \stringembed, and \affbertsim between \semanticembed and \stringembed. Specifically, the affinity matrices, \affgraphbert, \affgraphsim, and \affbertsim are calculated as follows:

\begin{equation}
\begin{aligned}
    A^{(hb)} &= \text{tanh}\big( \boldsymbol{X_{h}} W^{(hb)} \boldsymbol{X_{b}}^T \big) \\
    A^{(hs)}  &= \text{tanh}\big( \boldsymbol{X_{h}} W^{(hs)} \boldsymbol{X_{s}}^T \big) \\
    A^{(bs)}  &= \text{tanh}\big( \boldsymbol{X_{b}} W^{(bs)} \boldsymbol{X_{s}}^T \big)
\end{aligned}
\end{equation}
, where \weighttopobert, \weighttoposim, and \weightbertsim are learnable weights. Next, we measure the weight \attentiongraph for \topoembed, acknowledging supports from the other embedding views \semanticembed and \stringembed through \affgraphbert and \affgraphsim as follows:
\begin{equation}
    K_h = \text{tanh}
                    \big(
                        W_h\boldsymbol{X_{h}} + 
                        W_{bh} (A^{(hb)} \boldsymbol{X_{b}})   +
                        W_{sh} (A^{(hs)} \boldsymbol{X_{s}}) 
                    \big)
\end{equation}

In the same manner, we compute the weights \attentionbert for \bertembed, and \attentionstring for \stringembed as follows:
\begin{equation}
\begin{aligned}
    K_b = \text{tanh}
                    \big(
                        W_b \boldsymbol{X_{b}} + 
                        W_{hb} (A^{(hb)} \boldsymbol{X_{h}})   +
                        W_{sb} (A^{(bs)} \boldsymbol{X_{s}}) 
                    \big) \\
    K_s = \text{tanh}
                    \big(
                        W_s \boldsymbol{X_{s}} + 
                        W_{hs} (A^{(hs)} \boldsymbol{X_{h}})   +
                        W_{bs} (A^{(bs)} \boldsymbol{X_{b}}) 
                    \big) 
\end{aligned}
\end{equation}

Then, the co-attentive multi-aspect embeddings \outputattgraph, \outputattbert and \outputattstring can be computed as follows:
\begin{equation}
\label{eq:triplet-embeds}
\begin{aligned}
    \boldsymbol{\hat{X}_{h}} &= {softmax}(K_h) \odot \; \boldsymbol{X_{h}} \\
    \boldsymbol{\hat{X}_{b}} &= {softmax}(K_b) \odot \boldsymbol{X_{b}} \\
    \boldsymbol{\hat{X}_{s}} &= {softmax}(K_s) \odot \; \boldsymbol{X_{s}}
\end{aligned}
\end{equation}
where $\odot$ is the element-wise product.

\subsection{Reasoning-based Representations}
\begin{table*}[t]
\centering
  \caption{
  Neural Logical Regularizations. The NOT module is implemented by an one-layer MLP, and the OR module is implemented by another one-layer MLP. The \emph{True} and \emph{False} are logical constants in the traditional logical equations, but are learnable representations in our neural logical reasoning modules.
  }
  \resizebox{0.9\linewidth}{!}{
  \begin{tabular}{l|l|l|l}
  \toprule
       &       \multicolumn{1}{c|}{Logical Rule}    &   \multicolumn{1}{c|}{Equation} & \multicolumn{1}{|c}{Neural Logical Regularization.}\\
    \midrule
       \multirow{2}{*}{NOT} & Negation & $\neg True = False$ &  
                            $r_1=
                                \sum^{}_{j \in \mathcal{X}} sim(\job, NOT(\job)) +
                                \sum^{}_{\standardjob \in \mathcal{Y}} sim(\standardjob, NOT(\standardjob))
                            $ \\
                            & Double Negation & $\neg (\neg \job) = \job$ & 
                            $r_2 = 
                                \sum^{}_{j \in \mathcal{X}} \big(1 - sim(\job, NOT(NOT(\job))) \big)+ 
                                \sum^{}_{\standardjob \in \mathcal{Y}} \big(1 - sim(\standardjob, NOT(NOT(\standardjob))) \big)
                            $\\
    \midrule
    \multirow{4}{*}{OR} & Identity & $\job \; \vee \; False = \job$ & 
                            $r_3=
                                \sum^{}_{\job \in \mathcal{X}} \big(  1 - sim(OR(\job, False), \job) \big) +
                                \sum^{}_{\standardjob \in \mathcal{Y}} \big( 1 - sim(OR(\standardjob, False), \standardjob) \big)
                            $ 
                            \\
                       & Annihilator & $\job \; \vee \; True = True$ & 
                            $r_4=
                                \sum^{}_{\job \in \mathcal{X}} \big(  1 - sim(OR(\job, True), True) \big) +
                                \sum^{}_{\standardjob \in \mathcal{Y}} \big( 1 - sim(OR(\standardjob, True), True) \big)
                            $ 
                            \\
                     & Idempotence & $\job \; \vee \; \job = \job$ & 
                            $r_5=
                                \sum^{}_{\job \in \mathcal{X}} \big(  1 - sim(OR(\job, \job), \job) \big) +
                                \sum^{}_{\standardjob \in \mathcal{Y}} \big( 1 - sim(OR(\standardjob, \standardjob), \standardjob) \big)
                            $ 
                            \\
                     & Complementation & $\job \; \vee \; \neg\job = True$ & 
                            $r_6=
                                \sum^{}_{\job \in \mathcal{X}} \big(  1 - sim(OR(\job, NOT(\job)), True) \big) +
                                \sum^{}_{\standardjob \in \mathcal{Y}} \big( 1 - sim(OR(\standardjob, NOT(\standardjob)), True) \big)
                            $ 
                            \\
            
    \bottomrule
  \end{tabular}
  }
  \label{tab:logical-reg}
\end{table*}

Mapping an input job title $\job$ to a normalized job title $\standardjob$ based solely on their similarity score is unwary. To alleviate uncertainty issues, it is necessary to also consider the similarity scores of $\job$ with the rest of the normalized job titles in the set $\mathcal{Y}$. For example, if the mapping score between $\job$ and a certain $\standardjob_k \in \mathcal{Y}$ is high at $0.99$, while the mapping scores between $\job$ and the other $\standardjob_l \in \mathcal{Y}$ ($l \ne k$) are low at $0.01$, then it is considered \emph{certain} that $\job$ maps to $\standardjob_k$. However, if the mapping score between $\job$ and $\standardjob_k \in \mathcal{Y}$ is high at $0.9$, and the mapping scores between $\job$ and a few other $\standardjob_l \in \mathcal{Y}$ ($l \ne k$) are close to ($\job$, $\standardjob_k$), then there is high \emph{uncertainty} when mapping $\job$ to $\standardjob_k$, even though its mapping score is the highest. Therefore, we need a mechanism that takes into account mapping scores of $\job$ with all $\standardjob_k \in \mathcal{Y}$ simultaneously.

In other words, we need a mechanism that considers collaborative supports across all the mapping scores. Specifically, in the example above, the mapping decision can be made by a reasoning procedure that checks if $\job$ is mostly similar to $\standardjob_k$, and totally dissimilar to the rest of the job titles $\standardjob_l \in \mathcal{Y}$ ($l \ne k$), and concludes that $\job$ maps to $\standardjob_k$. Such a reasoning procedure can be represented as a logical structure, leading us to use neural collaborative reasoning \cite{chen2021neural}. Furthermore, our ablation study demonstrates that the reasoning improves the performance of {\jtm}. Thus, we can represent such a reasoning procedure as a logical structure, as shown below:
\begin{equation}
\label{eq:sample-reasoning}
sim(\job_i, \standardjob_1) \wedge sim(\job_i, \standardjob_2) \wedge {sim}(\job_i, \standardjob_3) \rightarrow \standardjob_3
\end{equation}

Hence, we are inspired to design a neural collaborative reasoning module \cite{chen2021neural} that learns reasoning-based representations of the input job titles. In this sense, the problem of predicting $\standardjob_2$ as a correct mapping or not with the example above (\ie, Equation~(\ref{eq:sample-reasoning})) is converted into the problem of deciding if the following Horn clause is True or False:
\begin{equation}
\label{eq:sample-horn}
sim(\job_i, \standardjob_1) \wedge sim(\job_i, \standardjob_2) \rightarrow {sim}(\job_i, \standardjob_3) 
\end{equation}

Note that due to the lack of topological information of normalized job titles, we are not able to produce topological embeddings for normalized job titles. However, producing semantic embeddings and syntactic embeddings for normalized job titles is straight-forward and follows a similar process as for input job titles. As such, we define a Horn clause for finding a mapping between the input job title $\job_i$ and a correct mapping standard job title $\standardjob_c \in \mathcal{Y}$ with regard to the input semantic embeddings of both $\job_i$ and $\standardjob_k$ can be defined as follows:
\begin{equation}
\label{eq:horn-bert}
\resizebox{0.9\linewidth}{!}{$
    sim(\job_i^{(b)}, \standardjob_1^{(b)}) \wedge \dots \wedge {sim}(\job_i^{(b)}, \standardjob_{|\mathcal{Y}|}^{(b)}) \rightarrow {sim}(\job_i^{(b)}, \standardjob_c^{(b)})  
$}
\end{equation}

Based on the De Morgan's Law, we can re-write Equation~(\ref{eq:horn-bert}) using only two basic logical operator \emph{OR} (\ie, $\vee$) and \emph{NOT} (\ie, $\neg$) and obtain the reasoning-based representation \outputreasonbert of $\job_i$ as follows: 
\begin{equation}
\label{eq:horn-bert-final1}
    \boldsymbol{{X}^\prime_{b}}  = \neg sim(\job_i^{(b)}, \standardjob_1^{(b)}) \vee \dots \vee \neg{sim}(\job_i^{(b)}, \standardjob_{|\mathcal{Y}|}^{(b)}) \vee {sim}(\job_i^{(b)}, \standardjob_c^{(b)})  
\end{equation}

Similarly, we can obtain the reasoning-based representation \outputreasonstring of $\job_i$ with regard to the syntactic embedding view as follows: 
\begin{equation}
\label{eq:horn-bert-final2}
    \boldsymbol{{X}^\prime_{s}}  = \neg sim(\job_i^{(s)}, \standardjob_1^{(s)}) \vee \dots \vee \neg{sim}(\job_i^{(s)}, \standardjob_{|\mathcal{Y}|}^{(s)}) \vee {sim}(\job_i^{(s)}, \standardjob_c^{(s)})  
\end{equation}

\begin{figure}[t]
  \centering
  \includegraphics[width=\linewidth]{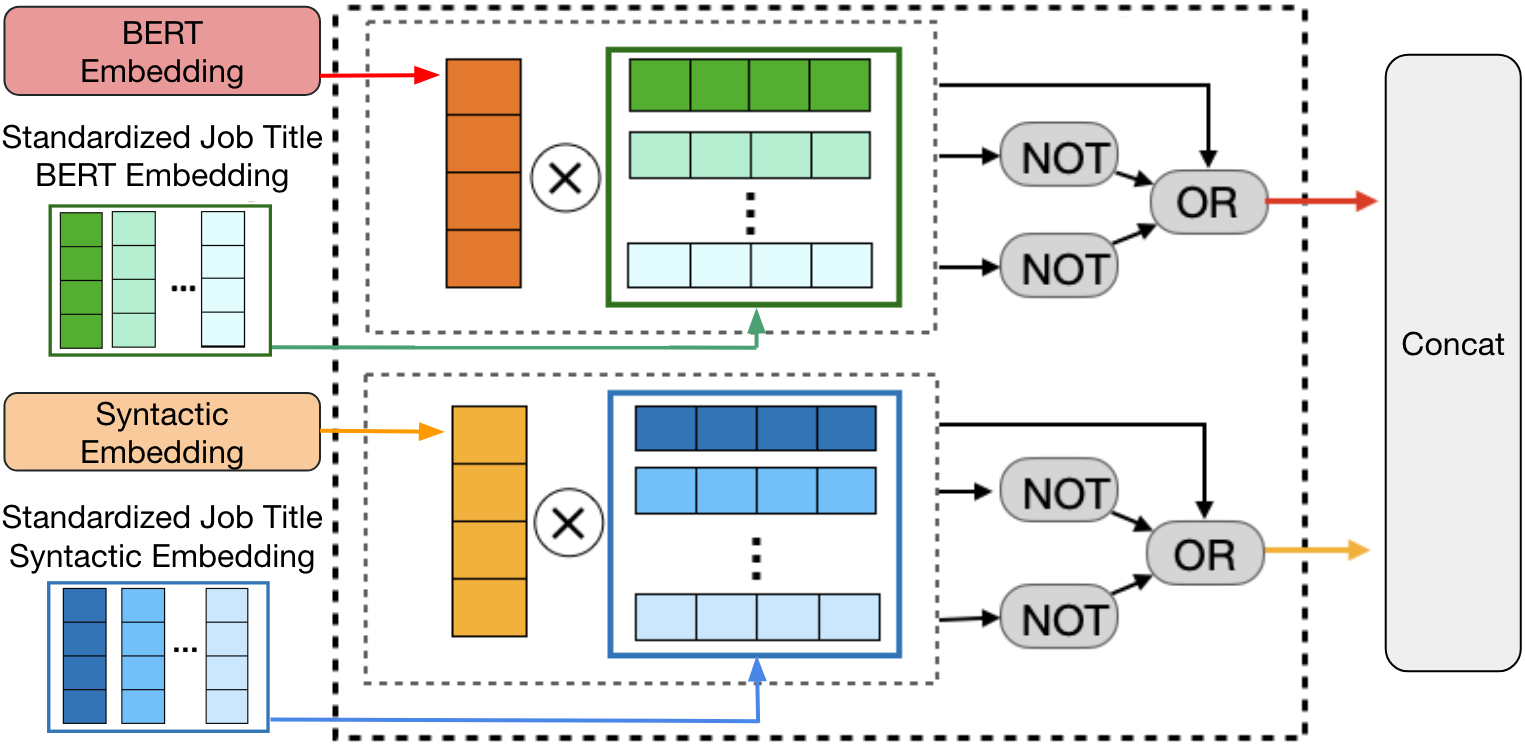}
  \caption{Architecture for reasoning-based representation.}
  \label{fig:reasoning}
  \vspace{-10pt}
\end{figure}

Figure \ref{fig:reasoning} summarizes our architecture for the neural logical reasoning. 
With reasoning-based representations \outputreasonbert and \outputreasonstring are now established together in Equation (\ref{eq:horn-bert-final1}) and (\ref{eq:horn-bert-final2}), as well as co-attentive multi-aspect embeddings \outputattgraph, \outputattbert, and \outputattstring (Equation (\ref{eq:triplet-embeds})), we next fuse these embeddings to have a final representation of the input job title.

\subsection{Fusion}
We concatenate the reasoning-based representations \outputreasonbert and \outputreasonstring, and the co-attentive multi-aspect embeddings \outputattgraph, \outputattbert, and \outputattstring. Then we project the final job title embeddings into the size of all standard job titles $\vert \mathcal{Y} \vert$ and generate a class probability distribution through the \textit{softmax} operator.
\begin{equation}
\resizebox{0.75\linewidth}{!}{
    $\hat{y} = softmax(ReLU(W($ [\outputattgraph; \outputattbert; \outputattstring; \outputreasonbert; \outputreasonstring] $)))$
}
\end{equation}

\subsection{Learning Objective}
We use the categorical cross-entropy as the loss function to train our {\ours}. The categorical cross-entropy loss function is defined as following:
\begin{equation}
    L(\theta) = -\sum_{\job \in {\mathcal{Y}}} y_j log(\hat{y}_j)
\end{equation}
where $\theta$ 
refers to all the parameters in the entire model. 

In our implementation for reasoning-based representaion (Equation (\ref{eq:horn-bert-final1}) and (\ref{eq:horn-bert-final2})), following \cite{chen2021neural}, the OR module is implemented by a multi-layer perceptron (MLP) with one hidden layer, and the NOT/NEGATION module is also implemented by another multi-layer perceptron. To explicitly guarantee that these OR and NOT modules implement the expected logic operations, we constraints them with logical regularization as defined in Table \ref{tab:logical-reg}. The final loss function of our {\ours} is defined as followings:
\begin{equation}
    L(\theta) = -\sum_{\job \in {\mathcal{Y}}} y_j log(\hat{y}_j) + \sum^{6}_{q=1} {r_q}
\end{equation}
where $\sum^{6}_{q=1} {r_q}$ is the summation of all six neural logical regularizations that are defined in Table \ref{tab:logical-reg}.

\section{Empirical Validation}
In this section, we present the evaluation results of our proposed {\ours} model against competing baselines. We use our large-scale {\jtm} dataset for the comparison, as other {\jtm} datasets from \cite{zhang2020large,dave2018combined,li2017nemo} are not publicly available. We attempt to answer the following Evaluation Questions (EQ):
\squishlist
\item \textbf{EQ1}: How does {\ours} perform against the baselines?
\item \textbf{EQ2}: Which components in {\ours} are more helpful?
\item \textbf{EQ3}: Can {\ours} be useful for other downstream tasks?
\squishend

\subsection{Experimental Settings}
\subsubsection{Baselines} 
We compare {\ours} against an exhaustive list of ten baseline models, including traditional simple solutions and state-of-the-art models: KNN-based \cite{zhu2016semantic}, Word2Vec-based \cite{boselli2017using}, DeepCarotene \cite{wang2019deepcarotene}, Node2Vec \cite{grover2016node2vec}, GloVe \cite{pennington2014glove}, NEO \cite{giabelli2020neo}, WoLMIS \cite{boselli2018wolmis}, SBERT \cite{reimers2019sentence}, Job2Vec \cite{zhang2019job2vec}, and Universal Sentence Encoder (USE) \cite{cer2018universal}. 
Note that as job descriptions are not available in our dataset, we construct the baseline models using only job titles to enable a fair comparison. 

\subsubsection{Evaluation protocols} 
To evaluate the performance of all compared models, we use two widely used ranking metrics, \emph{Precision}@N and \emph{NDCG}@N, with $N$ being the top-$N$ results produced by each model. \emph{Precision}@N accounts for the number of relevant results among top-$N$ output candidates, while \emph{NDCG}@N applies an increasing discount of \emph{log2} to items at lower ranks. We divide the dataset into 64\%, 16\%, and 20\%, where we train for 64\% using 16\% as a validation, and then test for 20\% for the {\jtm} task.
Regarding our implementation settings, we use their reported hyperparameter settings for baseline models. For the GloVe-based (word-based) models, the dimension is set to 300. 
For the Universal Sentence Encoder (USE), we use 512 dimensions, which is the default setting by the provider, and the SBERT's embedding size is set to 768 for the same reason. For Node2Vec, we choose 128 accounting for execution time on a large-scale graph dataset. For our model, we vary the embedding size from \{128, 256, 512\}. During training, the number of epochs is set to 200 with early stopping. Our model and all the baselines are trained with a batch size of 256 using the Adam optimizer and learning rate of $10^{-3}$.
\vspace{-2pt}

\begin{table}[t]
\centering
  \caption{Precision@10 and NDCG@10 of {\ours} and baseline models on our dataset. The best results are in \textbf{bold}, the best baseline's performance is \underline{underlined}.}
  \resizebox{\linewidth}{!}{
  \def\arraystretch{1.2}
  \begin{tabular}{p{0.5cm}|l|l|c|c}\hline
   & Model & Venue & Precision@10 & NDCG@10 \\\hline
    (\romannumeral 1) & KNN-based \cite{zhu2016semantic}            & CoRR'16   & 0.0913                & 0.0871 \\
    (\romannumeral 2) & Word2Vec-based \cite{boselli2017using}      & ECML'17   &0.1254                & 0.0544 \\
    (\romannumeral 3) & DeepCarotene \cite{wang2019deepcarotene}    & BigData'19&0.1255                & 0.0543 \\
    (\romannumeral 4) & Node2Vec \cite{grover2016node2vec}          & KDD'16    &0.1255                & 0.0609 \\
    (\romannumeral 5) & GloVe \cite{pennington2014glove}            & EMNLP'14  &0.3080                & 0.1817 \\
    (\romannumeral 6) & NEO \cite{giabelli2020neo}                  & ISWC'20   &0.3422                & 0.2054 \\
    (\romannumeral 7) & WoLMIS \cite{boselli2018wolmis}             & IIS'18    &0.3536                & 0.2480 \\
    (\romannumeral 8) & SBERT \cite{reimers2019sentence}    & EMNLP'19  &0.6121    & 0.4720 \\
    (\romannumeral 9) & Job2Vec \cite{zhang2019job2vec}             & CIKM'19   &0.6122                & 0.4622 \\
    (\romannumeral 10) & USE \cite{cer2018universal}                & EMNLP'18  &\underline{0.6619}                & \underline{0.4887} \\\hline
    & \textbf{{\ours}} & Ours    &\textbf{0.7285}       &\textbf{0.5743}\\\hline
  \end{tabular}
  \label{tab:classification}
  }
  \vspace{-10pt}
\end{table}

\subsection{EQ1: Performance of {\ours}}
Table \ref{tab:classification} shows the overall performance of {\ours} and the compared models on our large-scale {\jtm} dataset. We observe that word-based baselines (baseline (\romannumeral 1, \romannumeral 2, \romannumeral 3)) perform the worst. This can be attributed to two main reasons. First, word-level baselines mostly rely on word embedding techniques and do not account for contextual word semantic relationships in job titles, which results in a failure to mitigate the interdisciplinary correlation among job titles. Second, the word-level semantic baselines use additional job descriptions to enhance their performance, but job descriptions are not always publicly available in {\jtm} datasets, have limited access, and are expensive to collect. Although baseline (\romannumeral 5) performs better than the word-based baselines, its performance is still significantly lower than other models. {\ours} significantly outperforms Node2Vec (baseline (\romannumeral 4)), indicating that using only graph representation learning is suboptimal.

Models that are more applied and job-specific (baseline (\romannumeral 6, \romannumeral 7)) achieve higher performance compared to word-level semantic models and topological baseline, as they are able to deal with both messy and interdisciplinary job titles, though {\ours} still outperforms them. The sentence-level semantic-based models (baseline (\romannumeral 8, \romannumeral 9, \romannumeral 10)) perform very well in contrast to other baselines because they can extract and represent semantic meanings using pretrained models.

In short, {\ours} vastly outperforms all related baselines by utilizing our multi-aspect co-attentive reasoning representation. An important reason for this is that all prior models were developed using \emph{company-generated} ``pure" job titles, while our dataset is \emph{user-generated} ``impure" job titles. Compared to the best baseline, \ie, USE, {\ours} improves \emph{Precision@10} by 10.06\% and \emph{NDCG@10} by 17.52\%, confirming the effectiveness of {\ours}.

\subsection{EQ2: Ablation Study}
We conducted an ablation study to address EQ2. Since SBERT (\ie, baselines \romannumeral 8 \  and \romannumeral 10) yielded relatively good performance and BERT embedding is an essential feature of our model, we evaluated the performance of removing each single component of {\ours} except for the BERT embedding. Table \ref{tab:ablation} presents the results. We observed two main findings.
First, hyperbolic graph embeddings have a significant contribution to the {\jtm} performance of {\ours}. Removing this component from {\ours} without co-attention and reasoning led to a reduction of \emph{Precision@10} by 18.28\% and \emph{NDCG@10} by 19.62\%. This demonstrates the effectiveness of hyperbolic graph embeddings for the major issues in {\jtm} task, such as overlapping and messy job titles (i.e., Challenge 1-3 in Introduction). 
Second, co-attention (CoAtt) and reasoning-based representation (Reasoning) improved the performance from the simple concatenation model by 4.70\% in \emph{Precision@10} and 7.47\% in \emph{NDCG@10}, showing the effectiveness of multi-aspect co-attention and fusion of co-attentive multi-aspect embeddings and reasoning-based representations. In summary, the removal of each component in {\ours} reduced its performance, indicating the effectiveness of our model. 

\begin{table}[t]
\centering
  \caption{
  Ablation study experiments for {\ours}.
  }
  \resizebox{\linewidth}{!}{
  \begin{tabular}{l|l|l}
  \toprule
    Model   &       \multicolumn{1}{c|}{Precision@10}    &   \multicolumn{1}{c}{NDCG@10}\\
    \midrule
    {\ours} & \text{ }\textbf{0.7285}       &\text{ }\textbf{0.5743} \\
    \midrule
    - CoAtt ($ = \alpha$)   & \text{ }0.6996 ($\downarrow$ 4.13\%) \text{ }& \text{ }0.5630 ($\downarrow$ 2.01\%) \text{ }\\
    - $\alpha$ - Reasoning ($ = \beta$)  & \text{ }0.6958 ($\downarrow$  4.70\%) \text{ }& \text{ }0.5344 ($\downarrow$ 7.47\%) \text{ }\\
    - $\beta$ - Syntactic     & \text{ }0.6273 ($\downarrow$  16.13\%) \text{ }& \text{ }0.4859 ($\downarrow$ 18.19\%) \text{ }\\
    - $\beta$ - Hyperbolic Graph ($ = \kappa$)         & \text{ }0.6159 ($\downarrow$  18.28\%) \text{ }& \text{ }0.4801 ($\downarrow$ 19.62\%) \text{ }\\
    - $\kappa$ - Syntactic & \text{ }0.6121 ($\downarrow$  19.02\%) \text{ }& \text{ }0.4720 ($\downarrow$ 21.67\%) \text{ }\\
    \bottomrule
  \end{tabular}
  }
  \label{tab:ablation}
  \vspace{-10pt}
\end{table}

\subsection{EQ3: Other Downstream Tasks} 
To assess the performance of {\ours} in other job-domain downstream tasks, we conducted additional experiments as follows.

\subsubsection{Link Prediction} 
Link prediction is one of the most common tasks for graphs and networks \cite{kumar2020link, cai2020multi, zhang2018link}. In this part, to see the effectiveness of the multi-aspect embeddings learned by {\ours}, we present an additional capability of {\ours} in the link prediction task. We compare {\ours} with Node2Vec, Word2Vec, GloVe, USE, and Job2Vec as representative baselines. 

To generate the training/development/testing sets for the link prediction task, we randomly removed 20\% of the total number of links in the graph, considering them as the positive links in the testing set, and sampled the same amount of negative links in the graph for the testing set. We also randomly removed 20\% of the positive links in the remaining 80\% positive links and sampled the same amount of negative links to form a development set. The rest of the graph was kept as a training set. Then, we followed \cite{grover2016node2vec} and used different binary operators (\ie, Average, Hadmard, Weighted-L1, and Weighted-L2) to obtain the link embedding from the employee node embedding, the job-title node embedding obtained by {\ours}, and the link/edge that connects these two nodes. We selected the best binary operator using the development set and reported the performance metric using AUC.

Table \ref{tab:linkprediction} shows the performance of {\ours} and the compared methods. The best baseline is \textit{Job2Vec}, and we observe that {\ours} outperforms all of the baselines. Specifically, {\ours} relatively improves Job2Vec by 5.59\% of AUC. This result demonstrates that the multi-aspect embeddings from our {\ours} model are effective not only for {\jtm} task but also for link prediction.

\begin{table}[tb]
\centering
  \caption{AUC in link prediction task.}
  \resizebox{0.67\linewidth}{!}{
  \begin{tabular}{p{4cm}|c}\hline
  \toprule
    Method & AUC\\
    \midrule
    Word2Vec-based~\cite{boselli2017using}&0.5648\\
    GloVe~\cite{pennington2014glove}&0.6278\\
    USE~\cite{cer2018universal}&0.8370\\
    Node2Vec~\cite{grover2016node2vec}&0.8743\\
    Job2Vec~\cite{zhang2019job2vec}&0.9431\\
    \bottomrule
    {\ours} & \textbf{0.9957}\\\hline
  \end{tabular}
  }
  \label{tab:linkprediction}
\end{table}

\begin{table}[tb]
\centering
  \caption{Job Mobility Prediction}
  \resizebox{0.82\linewidth}{!}{
  \begin{tabular}{p{5cm}|c}\hline
  \toprule
    Method& MAP@10\\
    \midrule
    No {\jtm} (unpreprocessed) + NEMO~\cite{li2017nemo}&0.5349\\
    Job2Vec~\cite{zhang2019job2vec} + NEMO~\cite{li2017nemo}&0.6418\\
    USE~\cite{cer2018universal} + NEMO~\cite{li2017nemo}&0.6529\\
    \bottomrule
    {\ours} + NEMO~\cite{li2017nemo}& \textbf{0.7013}\\\hline
  \end{tabular}
  }
  \label{tab:jobmobility}
  \vspace{-10pt}
\end{table}


\subsubsection{Job Mobility Prediction.}
Job mobility prediction is an essential job-domain downstream task \cite{li2017nemo, meng2019hierarchical, zhang2021attentive, wang2021variable}. It involves predicting a user's next job titles based on their sequence of job trajectories, wherein {\jtm} is conducted for preprocessing the dataset. 
Specifically, we use {\ours} to preprocess our resume dataset, by converting each input job title with the \emph{top-1} matching standard job title. To compare with our {\ours}, we use Job2Vec, USE as baselines. We also prepare the unpreprocessed dataset. 
For the job mobility prediction model, we adopt NEMO \cite{li2017nemo} as it is the state-of-the-art model, and evaluate its performance using mean average precision at 10 (MAP@10) as the metric. Note that we only consider the job title transition information for our evaluation, as other features are costly to collect and are not available in our dataset. We compare the impact of the {\jtm} methods on the model's performance. 

Table \ref{tab:jobmobility} shows the performance of the job mobility prediction task using the {\ours} and baselines for preprocessing. The results demonstrate that the {\ours} has a considerable impact on the job mobility prediction model's performance. Its multifaceted mapping approach assists job mobility prediction models in learning more effectively, resulting in a performance improvement. Although the accuracy improvement alone does not provide insights into user behavior, the effectiveness of {\ours} in improving performance suggests its potential applicability in various job-domain downstream tasks.

\section{{\ours} API and Use Cases}
\begin{figure}[t]
    \centering
    \fbox{\includegraphics[width=0.90\linewidth]{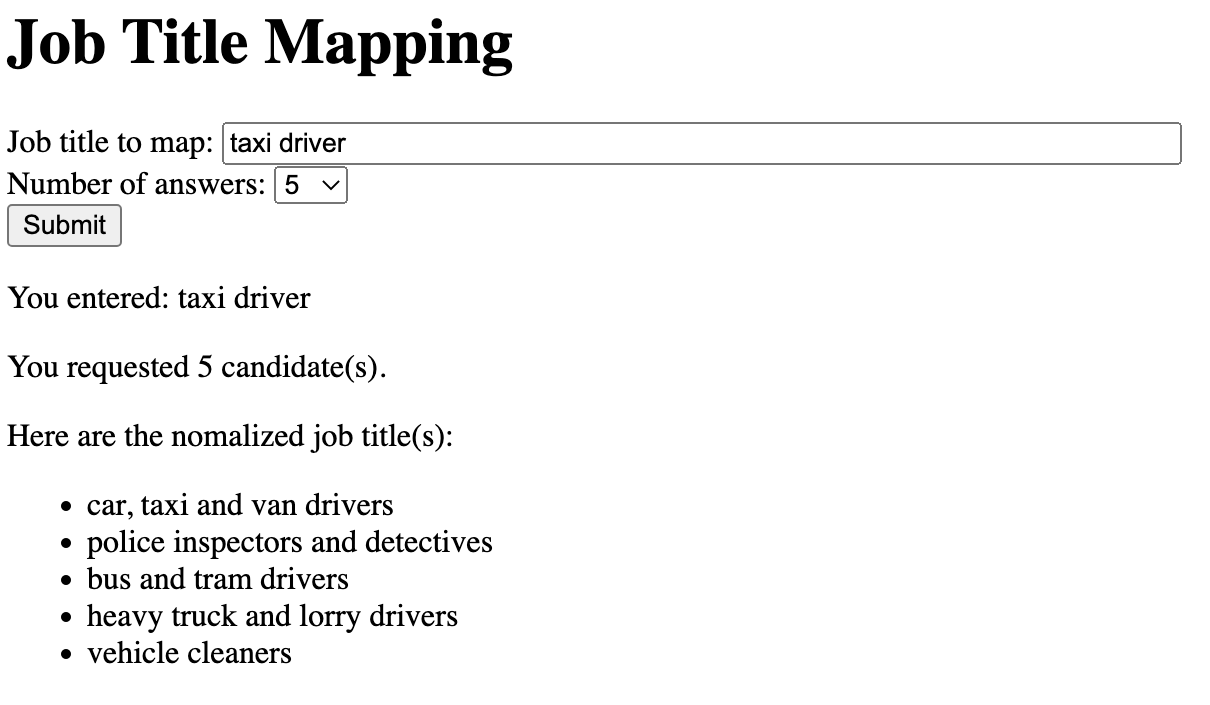}}
        \caption{{\ours} API demo page. This shows an example of the job title normalization for ``taxi driver''}
    \label{fig:API_page}
    \vspace{-10pt}
\end{figure}

To make {\ours} accessible as a public resource, we release a RESTful API for {\ours}. This API allows users to input any textual job title entity and receive the corresponding normalized job titles based on the public taxonomy (\ie, ESCO). Users are able to obtain up to the top-5 most relevant normalized job titles, which helps individuals or organizations in preprocessing and cleansing their job title datasets for job-domain downstream tasks. The API provides the output predicted by {\ours} if the input job entity is included in our graph. Otherwise, the API employs textual embeddings to get the output so that users can use any text and retrieve relevant normalized job titles. 
We also built a demo website that users can touch and see the results more intuitively. Figure \ref{fig:API_page} shows the screenshot of {\ours} web app that uses our API behind. 


Our {\ours} API has the potential for a wide range of applications and use cases, as listed below: 
\begin{itemize}
    \item \emph{Recruitment platforms}: By integrating job titles, {\ours} API enables recruiters and job seekers to more easily compare job titles and requirements across different companies and industries.
    \item \emph{Resume standardization}: {\ours} API can be incorporated to automatically normalize job titles in users-uploaded resumes on online platforms, facilitating the matchmaking between job seekers with job postings. 
    \item \emph{Market research}: {\ours} API is also beneficial in market and economic research for tracking job trends and analyzing job requirements across different industries and regions, as economic researchers typically need to clean job titles for their analysis. 
    \item \emph{Search query expansion}: By mapping all variations of an entity to a single normalized form, {\ours} API can be used to improve the relevance of job search results, expanding the search to include all documents that mention the normalized job titles.
\end{itemize}

\section{Conclusion}
In this paper, we proposed a novel job title normalization model, {\ours}, toward creating a fine-grained job taxonomy using a real-world and large-scale career trajectory dataset. Our approach utilized multi-aspect embeddings (\ie, graph, semantic, and syntactic embedding), multi-aspect co-attention, and reasoning-based representation to address the challenges of the \emph{Job Title Normalization} ({\jtm}) task effectively. 
We conducted extensive experiments comparing {\ours} to ten baseline models on the {\jtm} task and performed an ablation study. Furthermore, we conducted additional experiments on practical downstream tasks, such as link prediction and job mobility prediction, to assess the practical impact of our approach. Our results showed that: (1) {\ours} outperformed all baseline models in the {\jtm} task, and (2) {\ours} was effective in job-domain downstream tasks. Finally, we release {\ours} as an API for public use, which is useful for job-domain downstream tasks. 
\vspace{-2pt}


\section*{Acknowledgment}
This work was in part supported by NSF awards \#1934782 and \#1909702, and PSU CSRAI seed grant 2021.
\vspace{-2pt}

\bibliographystyle{IEEEtran}
\bibliography{dsaa23}
\end{document}